\documentclass[runningheads]{llncs}
\usepackage[T1]{fontenc}
\usepackage{graphicx}
\usepackage{enumitem}
\usepackage{amsmath}
\usepackage{amssymb}
\usepackage{algorithm}
\usepackage{algorithmicx}
\usepackage{algpseudocode}
\usepackage{bbm}   
\usepackage{booktabs}
\usepackage{threeparttable}
\usepackage{makecell} 
\usepackage{multirow} 
\usepackage{array} 
\usepackage{hyperref}
\usepackage{subcaption}
\usepackage{dblfloatfix} 
\usepackage{placeins}      
\usepackage{dblfloatfix} 
\usepackage{float}
\usepackage[table,xcdraw]{xcolor}

\begin{document}
\title{PECKER: A Precisely Efficient Critical Knowledge Erasure Recipe For Machine Unlearning in Diffusion Models}
\titlerunning{PECKER}
%
\author{Zhiyong Ma\inst{1,2} \and
Zhitao Deng\inst{1,3}\textsuperscript{*} \and 
Huan Tang\inst{1,3}\textsuperscript{*} \and 
Jialin Chen\inst{1} \and  
Zhijun Zheng\inst{3} \and 
Zhengping Li\inst{4} \and 
Qingyuan Chuai\inst{1}\textsuperscript{\#} 
}
\authorrunning{Z. Ma et al.}
%
\institute{Cao Tu Li(Guangzhou) Technology Co., Ltd, China \and
South China University of Technology, China \and
Guangzhou Xinhua University, China \and
Hong Kong Baptist University, HongKong \\
\email{\{* For equal contribution \& Corresponding Email: qingyuanchuai@ln.hk\}}
}
\maketitle              
\begin{abstract}
Machine unlearning (MU) has become a critical technique for GenAI models' safe and compliant operation.
While existing MU methods are effective, most impose prohibitive training time and computational overhead. 
Our analysis suggests the root cause lies in poorly directed gradient updates, which reduce training efficiency and destabilize convergence. To mitigate these issues, we propose PECKER, an efficient MU approach that matches or outperforms prevailing methods. Within a distillation framework, PECKER introduces a saliency mask to prioritize updates to parameters that contribute most to forgetting the targeted data, thereby reducing unnecessary gradient computation and shortening overall training time without sacrificing unlearning efficacy. Our method generates samples that unlearn related class or concept more quickly, while closely aligning with the true image distribution on CIFAR-10 and STL-10 datasets, achieving shorter training times for both class forgetting and concept forgetting. Our code is available at \url{https://github.com/zark-cyber/PECKER}.
\end{abstract}


\noindent\textbf{\noindent Warning:~This paper includes explicit visual content, discussions on adult material, racially sensitive language, and other topics that may be unsettling, distressing, or offensive to some readers.}

\section{Introduction}
Diffusion models have emerged as the backbone of real-world generative AI applications~\cite{Li2025Efficient,li2025preference,li2026sepprune}, spanning high-fidelity image synthesis and text-to-image systems~\cite{li2024nas,li2024pruning,li2026comprehensive}. 
As these models are deployed at scale in consumer platforms and enterprise workflows, they increasingly face stringent regulatory scrutiny (e.g., GDPR’s ``right to be forgotten``~\cite{GDPR}) and privacy risks, including unintended generation of sensitive personal data or non-compliant content~\cite{kong2025,lu2025hdcompression}.

A non-negotiable requirement for safe and compliant deployment is the ability to eliminate the influence of specific data or semantic categories (e.g.,a celebrity’s likeness or prohibited content) from pre-trained models~\cite{SFD,ESD}. 
While full retraining from scratch is theoretically viable for data deletion~\cite{SA,SLD}, it is computationally prohibitive.
This creates the need for efficient elimination of targeted semantics. 
To bridge this gap, Machine Unlearning (MU) has emerged as a promising paradigm, aiming to modify pre-trained models such that their behavior approximates a counterfactual model never exposed to the ``forget`` data~\cite{Bau,MU_Survey}.

\begin{figure*}[t]
    \centering
    \includegraphics[width=0.8\textwidth]{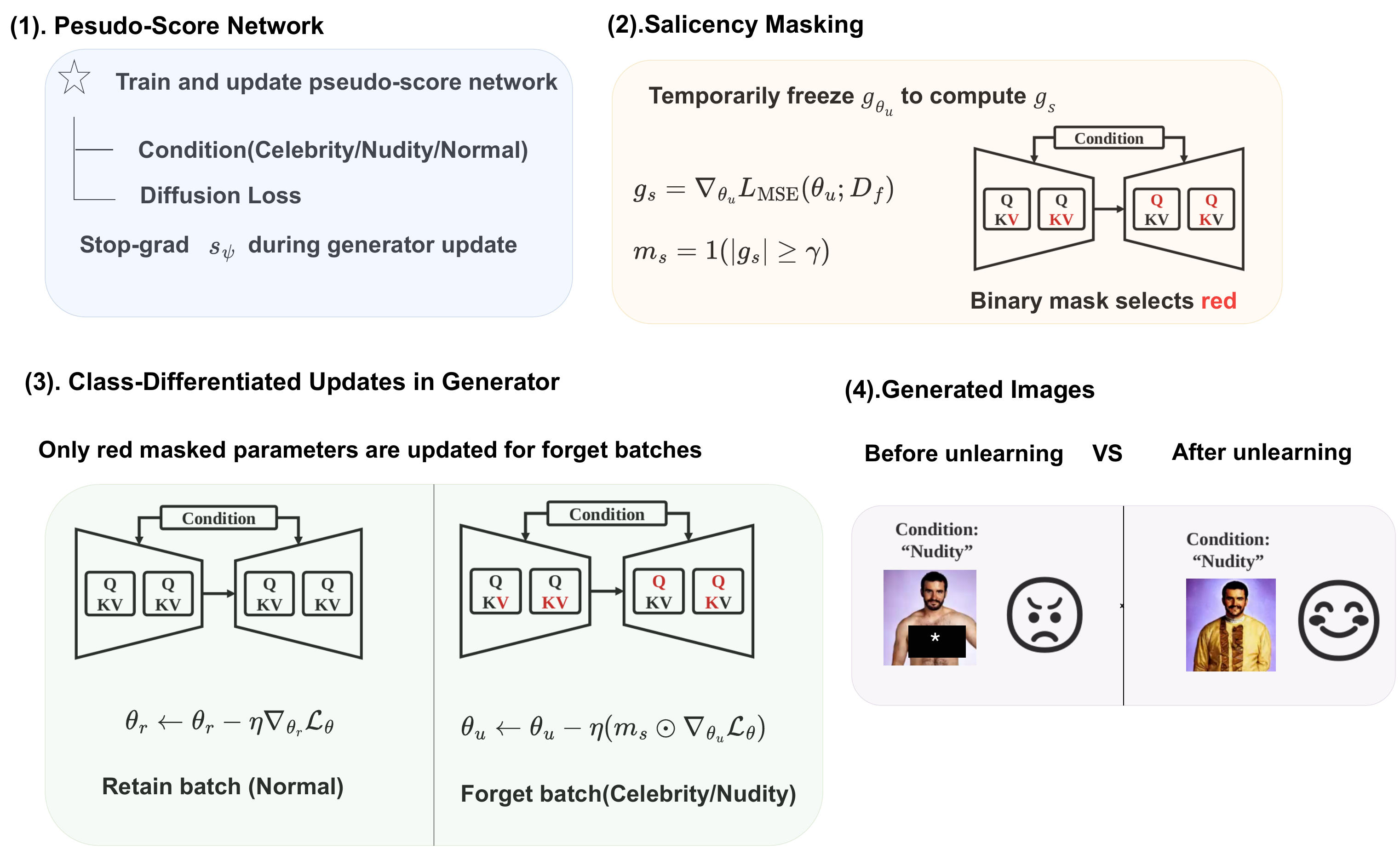}
    \caption{\textbf{Overview of PECKER.}
    (1) A pseudo-score network $s_\psi$ provides data-free supervision; gradients are stopped through $s_\psi$ during generator updates. (2) We compute saliency $g_s$ and a mask $m_s$ from a temporarily frozen generator (Eq. \ref{eq:gradient}–\ref{eq:saliency_mask}), where red marks masked parameters. (3) Retain batches use full updates, while forget batches update only masked parameters. (4) The resulting generator produces concept-erased outputs for the forget condition.}
    \label{fig:diagram}
\end{figure*}

However, applying MU to diffusion models remains challenging due to two inherent characteristics: (1) distributed semantic representation: meaningful concepts (e.g., ``cat``, ``celebrity face``) are encoded across deep, interconnected network layers rather than localized in specific parameters~\cite{DDPM,SD_v1.4}; (2) noise-predic\-tion objective: diffusion models optimize for denoising noisy inputs and thus make it difficult to isolate gradients related to specific target semantics~\cite{DDPM,diffusion_survey}.~T\-hese properties make it difficult to localize and suppress gradients tied to a specific concept.
Existing MU methods for diffusion models~\cite{ESD,SA} suffer from two key limitations: (a) naive parameter updates degrade performance on retained data; (b) indiscriminate weight tuning incurs excessive computational overhead. 
Both fail to address the distributed semantics and noise-prediction constraints of diffusion models~\cite{MU_Survey}.
Data-free behavior distillation approaches~\cite{li2025frequency,Zhou} provide privacy-friendly supervision via pseudo-samples and label-shifted targets, but lack saliency-aware parameter selection, often causing unfocused updates that harm non-target capabilities and slow convergence~\cite{SalUn,SFD}.

To resolve these limitations, we propose \textbf{PECKER} (Precisely Efficient Critical Knowledge Erasure Recipe), a saliency-scored distillation framework tailored for MU in diffusion models. 
PECKER couples behavior-level guidance with structure-aware parameter selection.
It comprises three mutually reinforcing components that enable precise and efficient erasure of critical knowledge:
\begin{itemize}[label=\textbullet, leftmargin=*]
  \item \textbf{Alternating Pseudo-Score Supervision.} We alternately update a lightwei\-ght pseudo score network $s_\phi$ to provide data-free, class-shifted supervision.Wh\-en updating the generator, we stop gradients through $s_\phi$.
  \item \textbf{Saliency Scoring and Masking.} We temporarily freeze the generator $g_{\theta}$ compute gradient saliency under the standard noise-prediction objective and derive a binary mask that pinpoints parameters encoding critical “forget” semantics.
  \item \textbf{Class-differentiated Updates in Generator Network.} For forget batches, we update only masked parameters to tighten the erasure region and accelerate convergence. For retain classes, standard full-parameter updates preserve utility.
\end{itemize}


Algorithm \ref{alg:pecker} summarizes PECKER’s iterative procedure, from pseudo-sample generation to class-differentiated parameter updates, implementing our key idea: focusing updates on salient parameters for forget classes while preserving utility on retain classes.

This design retains the privacy and deployability advantages of data-free distillation while improving unlearning specificity, efficiency, and stability.~Our main contributions are:
\begin{itemize}[label=\textbullet, leftmargin=*]
  \item We introduce a saliency-scored distillation framework that integrates parameter saliency with data-free distillation for diffusion-model unlearning.
  \item We propose a class-differentiated update rule: masked updates for forget classes and full updates for retain classes, balancing forgetting efficacy and retained utility.
  \item We demonstrate effectiveness across both class forgetting (CIFAR-10, STL-10) and concept forgetting (e.g., celebrity/nudity removal) tasks.
\end{itemize}


\section{Related Work}

\textbf{Unlearning in image generation models.} Post-hoc unlearning and model editing have attracted growing interest for privacy-preserving generation, motivated by regulations such as the EU’s GDPR~\cite{GDPR}. Early efforts primarily studied GANs~\cite{GANs}, where weight editing and data redaction were used to remove specific semantics (e.g., watermark removal)~\cite{Bau,Meng2022Locating}. However, GAN-specific edits often introduce artifacts at scale and do not readily transfer to diffusion models~\cite{SD_v1.4}. As diffusion models have become the dominant paradigm for high-fidelity generation, diffusion-oriented unlearning has received increasing attention.

\noindent\textbf{Concept forgetting in text-to-image diffusion models.}~Large text-to-image diffusion models~\cite{SD_v1.4,2,Hierarchical} raise safety concerns due to their ability to synthesize harmful or disallowed content, motivating concept erasure methods. Safe Latent Diffusion (SLD)~\cite{SLD,Safelatentdiffusion} performs inference-time steering via energy-based composition with classifier-free guidance (CFG)~\cite{CFG}. Erasing Stable Diffusion (ESD)~\cite{ESD} extends erasure to training time but still relies on CFG, limiting applicability to non-CFG diffusion models. Moreover, these approaches provide limited support for controlled concept replacement (i.e., redirecting a forgotten concept to a user-specified alternative)~\cite{SFD} and can be inefficient due to unfocused parameter updates~\cite{SalUn}.

\noindent\textbf{Saliency analysis in data and model parameters.} Prior saliency research mainly focuses on input-level explanations (saliency maps)~\cite{Simonyan,Zeiler,Smilkov} and data attribution~\cite{Zhou2016Learning,Chattopadhay2018GradCAM,li2023less}. In contrast, parameter-level saliency is less explored; pruning and sparsity methods~\cite{Han2016DeepCompression,Frankle2018LotteryTicket} can be viewed as identifying influential weights for efficiency. More recently, model editing in NLP ~\cite{Dai2022KnowledgeNeurons,DeCao2021Editing,Patil2024SensitiveInformation} studies localized “editable regions” that encode specific knowledge, aligning with the goal of identifying and modifying critical parameters for targeted unlearning.

\section{Preliminaries}

Before addressing the targeted  MU problem for conditional diffusion models, we first formalize core concepts and notations of diffusion modeling: A conditional diffusion model corrupts data $x_0 \sim p_{\text{data}}(x|c)$ during the forward diffusion process at step $t$ to yield a noisy observation $z_t = a_t x_{t-1} + \sigma_t \epsilon_t$, where $\epsilon_t \sim \mathcal{N}(0, \mathbf{I})$, $c$ denotes a guiding condition (e.g., class label or text prompt), and $a_t, \sigma_t$ are pre-defined diffusion scheduling parameters. For simplicity, we use the cumulative product $\alpha_t = \prod_{s=1}^t a_s$ (with $\sigma_t^2 = 1 - \alpha_t^2$) to express $z_t$ in a closed form relative to the clean data: $z_t = \alpha_t x_0 + \sigma_t \epsilon$ for $\epsilon \sim \mathcal{N}(0, \mathbf{I})$.

The pretraining objective of a diffusion model is to learn two key estimators: an optimal score estimator $s_\phi(z_t, c, t) = \nabla_{z_t} \ln p_{\text{data}}(z_t | c)$ (quantifying the gradient of the noisy data distribution’s log-probability), and an optimal conditional mean estimator $x_\phi(z_t, c, t) = \mathbb{E}[x_0 | z_t, c, t]$ (the most likely clean sample corresponding to $z_t$ under condition $c$). Leveraging Tweedie’s formula adapted to diffusion modeling, these two estimators are mutually derivable:
\begin{equation}
s_\phi(z_t, c, t) = \frac{a_t x_\phi(z_t, c, t) - z_t}{\sigma_t^2}, \quad x_\phi(z_t, c, t) = \frac{z_t + \sigma_t^2 s_\phi(z_t, c, t)}{a_t}.
\label{eq:equation_1}
\end{equation}
Equation (\ref{eq:equation_1}) shows that the score estimator ($s_\phi$) and conditional mean estimator ($x_\phi$) can be converted to each other, which is the theoretical basis for our one-step diffusion generator design.
With the learned score estimator, a reverse diffusion process can be constructed to sample from the target distribution via numerical discretization over the diffusion time horizon.

We focus on a distilled one-step diffusion generator $g_\theta(n, c)$ (where $n \sim \mathcal{N}(0, \mathbf{I})$), which produces samples from the pretrained model’s distribution in a single step. Let $\mathcal{D}_{\theta,c}$ denote the generative distribution of $x$ under condition $c$ via $g_\theta$, and $s_{\psi(\theta)}(z_t, c, t)$ be the score estimator corresponding to $g_\theta$. 
Analogous to the relation between $s_\phi$ and $x_\phi$, we have:
\begin{equation}
s_{\psi(\theta)}(z_t, c, t) = \frac{a_t x_{\psi(\theta)}(z_t, c, t) - z_t}{\sigma_t^2},
\end{equation}
where $x_{\psi(\theta)}(z_t, c, t)$ is the conditional mean estimator of the one-step generator.

For class-targeted unlearning in conditional diffusion models, our goal is to eliminate the model’s ability to generate samples for a specific ``forget class`` $c_f$, while preserving performance on all ``retain classes`` $C_r = \{c_r \mid c_r \neq c_f\}$. 
To achieve this, we replace $c_f$ with a ``cover class`` $c_0 \in C_r$, and align the generator’s distribution for $c_f$ with the original data distribution of $c_0$. 
Let $\mathcal{D}_r$ denotes the distribution of retain classes, $\mathcal{D}_s$ denotes the post-unlearning sampling distribution of all classes, and $\mathcal{D}_{\theta,c} = g_\theta(\mathcal{N}(0, \mathbf{I}), c)$ denotes the generator’s conditional distribution for class $c$. 
Our unlearning objective is twofold: (1) Align $\mathcal{D}_{\theta,c_f} \stackrel{d}{=} p_{\text{data}}(x | c_0)$ to erase $c_f$; (2) Ensure $\mathcal{D}_{\theta,c_r} \stackrel{d}{=} p_{\text{data}}(x | c_r)$ for all $c_r \in C_r$ to preserve retain class performance.

For concept-level unlearning (e.g., in text-to-image diffusion models), our goal extends to removing associations with specific keywords or named entities by substituting them with generic descriptions (e.g., ``a middle-aged man or women``). 
This process prioritizes minimizing degradation to the generation quality of other concepts, thus maintaining the model’s overall diversity and integrity under textual guidance.

\section{Method}
\label{sec:Method}

Existing data-free MU methods typically utilize a pre-trained generator to synthesize pseudo-samples and employ a class replacement strategy to generate pseudo-supervision signals that guide the model to progressively diminish its response to the forgotten classes. 
Applying the framework of Score Forgetting Distillation~\cite{SFD}, we propose PECKER, which consists of a score network and a generator network, as shown in Fig.~\ref{fig:diagram}.
\begin{algorithm}[htbp]
\caption{PECKER: Precisely Efficient Critical Knowledge Erasure Recipe}
\label{alg:pecker}
\begin{algorithmic}[1]
\Require
pre-trained score network $s_\phi$, generator $g_\theta$, fake score network $s_\psi$; hyperparameter $\xi$; label/concept to forget $c_f$, cover/override label $c_o$; coefficients $\lambda_\psi, \mu_\psi$ (for $\psi$), $\lambda_\theta, \mu_\theta$ (for $\theta$); $t_{\min} < t_{\text{init}} \le t_{\max}$; learning rate $\eta$
\State Initialization: $\theta_u \gets \phi$, $\theta_r \gets \phi$, $\psi \gets \phi$
\Repeat
\State Sample $c_r \sim \mathcal{D}_r$, $n_r, n_f \sim \mathcal{N}(0, \mathbf{I})$; 
\State $x_r \gets g_\theta(\sigma_{\text{init}} n_r, c_r, t_{\text{init}})$, $x_f \gets g_\theta(\sigma_{\text{init}} n_f, c_f, t_{\text{init}})$
\State Sample $\epsilon_r, \epsilon_f \sim \mathcal{N}(0, \mathbf{I})$, $s, t \sim \text{Unif}[t_{\min}, t_{\max}]$;
\State $z_r \gets \alpha_s x_r + \sigma_s \epsilon_r$, $z_f \gets \alpha_t x_f + \sigma_t \epsilon_f$;
\State \textbf{Score update:}
\State $\mathcal{L}_\psi \gets \lambda_\psi \gamma(s) \| x_\psi(z_r, c_r, s) - x_r \|_2^2 + \mu_\psi \omega_t \| x_\psi(z_f, c_f, t) - x_f \|_2^2$
\State $\psi \gets \psi - \eta \nabla_\psi \mathcal{L}_\psi$
\State \textbf{Generator update:}
\State $\mathcal{L}_\theta \gets \lambda_\theta \hat{\mathcal{L}}_{\text{sfd}}(\theta, \psi; \phi, c_r, c_r, \xi) + \mu_\theta \hat{\mathcal{L}}_{\text{sfd}}(\theta, \psi; \phi, c_o, c_f, \xi)$
\State \textbf{Forgotten class ($c_f$):} $g_s \gets \nabla_{\theta_u} \mathcal{L}_{\text{MSE}}(\theta_u; \mathcal{D}_f)$; build saliency mask $m_s$
\State $\theta_u \gets \theta_u - \eta (m_s \odot \nabla_{\theta_u} \mathcal{L}_\theta)$
\State \textbf{Retained class ($c_r$):} $\theta_r \gets \theta_r - \eta \nabla_{\theta_r} \mathcal{L}_\theta$
\Until{maximum training steps or images seen is reached}
\State \textbf{Output:} updated generator $g_{\theta_u}$
\end{algorithmic}
\end{algorithm}
In PECKER, the loss function for the fake score in \textbf{the score network} can be formulated as:
\begin{equation}
\mathcal{L}_\psi = \underbrace{\lambda_\psi \gamma(s) \left\| x_\psi(z_r, c_r, s) - x_r \right\|_2^2}_{for~Remaining~Classes} + \underbrace{\mu_\psi \omega_t \left\| x_\psi(z_f, c_f, t) - x_f \right\|_2^2}_{for~Forgetting~Classes},
\label{eq:psi_loss}
\end{equation}
where the former term represents the diffusion loss for the remaining classes, while the latter denotes the equivalent loss for the forgetting classes.
The loss function for \textbf{the generated network} can be formulated as:
\begin{equation}
\mathcal{L}_{\theta} = \lambda_{\theta} \underbrace{\hat{\mathcal{L}}_{\text{sfd}}(\theta, \psi; \phi, c_r, c_r, \xi)}_{Distillation~Loss} + \mu_{\theta}\underbrace{\hat{\mathcal{L}}_{\text{sfd}}(\theta, \psi; \phi, c_0, c_f, \xi)}_{Forgetting~Loss},
\label{eq:theta_loss}
\end{equation}
which is a combination of distillation loss and forgetting loss:
\begin{equation}
\hat{\mathcal{L}}_{\text{sfd}}(\theta,\psi;\phi, c_{r}, c_{r}, \xi), \quad \text{where } z_{t},t,x \sim D_{\theta,c_{r}},
\label{eq:distill_loss_retain}
\end{equation}
\begin{equation}
\hat{\mathcal{L}}_{\text{sfd}}(\theta,\psi;\phi, c_{o}, c_{f}, \xi), \quad \text{where } z_{t},t,x \sim D_{\theta,c_{f}},
\label{eq:distill_loss_forget}
\end{equation}
\begin{equation}
\hat{\mathcal{L}}_{\text{sfd}}(\theta,\psi;\phi, c_{\dagger}, c_{\ddagger},\xi) = (1 - \xi) \omega_t \frac{a^2_t}{\sigma^4_t} \left\| Q \right\|^2 + \omega_t \frac{a^2_t}{\sigma^4_t} \left( Q \right)^T \left( x_\psi(z_t, c_\ddagger, t) - x \right),
\label{eq:distill_loss_detail}
\end{equation}
where $\xi \geq 0$ is a hyperparameter that typically set as 1 or 1.2. 
We follow~\cite{Zhou} and~\cite{Yin} to define $\omega_{t} = \frac{\sigma^4_t}{a^2_t} \frac{C}{\|x_{\phi}(z_{t},c,t)-x\|_{1,\text{sg}}}$, where $C$ is the data dimension and $\text{``sg``}$ stands for stop gradient.
And we denote $x_\phi(z_t, c_{\dagger}, t) -  x_\psi(z_t, c_{\ddagger}, t)$ as $Q$.

The purpose of the distillation loss is to keep the labels of the retained classes close to the labels of the retained classes, while the goal of the forgetting loss is to make the labels of the forgotten classes approach the labels of the covered classes (which belong to the retained classes), thereby achieving the specific forgetting effect.
Such methods achieve label-level shifts in the ``behavioral guidance`` aspect, causing the labels of the forgotten classes to move closer to the labels of the retained classes, and the labels of the retained classes to be closer to each other. This effectively performs the forgetting and retaining functions without the need to access the original data, making significant progress in ensuring user privacy.

However, a limitation is that during the gradient update process, the method fails to focus on the parameter regions that significantly contribute to the forgetting class information, leading to unclear gradient optimization directions. This results in low training efficiency, unstable convergence, and other computational cost issues.
Inspired by the significance parameters in SalUn~\cite{SalUn}, we attempt to incorporate significance parameters into the generator network of SFD framework \cite{SFD}, providing the model with clear optimization directions during updates. This accelerates gradient updates, reduces training time, improves training efficiency, and ensures that the model maintains good performance on retained classes. Thus, we build an MSE loss function for the dataset $D_{f}$ (intended to be forgotten) and take gradient for the loss function update:
\begin{equation}
\ell_{\text{MSE}}(\theta; D_f) = \mathbb{E}_{t, \epsilon \sim \mathcal{N}(0, 1)} \left[ \left\| \epsilon - \epsilon_\theta(x_t | c) \right\|_2^2 \right]
\label{eq:mse_loss}
\end{equation}
\begin{equation}
g_s = \nabla_\theta \mathcal{L}_{\text{MSE}}(\theta; D_f)
\label{eq:gradient}
\end{equation}

Subsequently, a saliency mask is created based on the update gradient to identify the parameter regions that significantly contribute to the forgotten class information (with contributions greater than or equal to a given threshold). These regions are prioritized for optimization during updates:
\begin{equation}
m_s = \mathbbm{1}(|g_s| \geq \gamma)
\label{eq:saliency_mask}
\end{equation}

Finally, the gradient $g$ is element-wise multiplied with the mask $m_{s}$, allowing us to perform gradient descent on the parameters where the mask is $\mathbbm{1}$—thereby providing clear optimization directions.

\section{Experiments}
\label{sec:Experiments}
\subsection{Implementation details}
In our experiments, we perform an evaluation of class forgetting in the Denoising Diffusion Probabilistic Model (DDPM)~\cite{DDPM}, which has been pre-trained on the widely-used CIFAR-10 and STL-10 datasets, frequently employed in previous research for evaluating memory unlearning (MU). For all methods, class 0 was selected as the target class for forgetting. Additionally, we evaluate our approach on concept forgetting tasks within text-to-image diffusion models using Stable Diffusion (SD)~\cite{SD_v1.4}, specifically focusing on celebrity and nudity forgetting.

For the DDPM baselines, we utilize the default 1000-step DDPM samplers to compute the Fréchet Inception Distances (FID) for the remaining classes. In the case of the SD baselines, we employ 50-step DDIM samplers for model generation. All experiments were conducted on two NVIDIA A100 GPUs.
For DDPM, all methods utilize the Adam optimizer to update the model parameters, with the learning rate $\eta$ for most baselines set to $1 \times 10^{-4}$. 
For SFD, $\eta$ is set to $1 \times 10^{-5}$ following its original settings.
For PECKER, $\eta$ is $1 \times 10^{-5}$. 
The batch size for CIFAR-10 and STL-10 are set to 128 and 64, respectively.
And in Algorithm \ref{alg:pecker}, we also set coefficients~$\lambda_\psi = \mu_\psi = \lambda_\theta = \mu_\theta = 1.0$.
\subsection{Evaluation}

\begin{table*}[htbp]
  \centering
  \small
  \begin{tabular}{|c|c|c|c|c|c|c|c|}
    \hline
    \textbf{Dataset} & \textbf{Method} & \textbf{UA($\uparrow$)} & \textbf{FID($\downarrow$)} & \textbf{IS($\uparrow$)} & \textbf{Precision($\uparrow$)}  & \textbf{Data-free} \\
    \hline
    \multirow{5}{*}{CIFAR-10}
    & Retrain & \underline{99.94} & 303.34 & 1.58 & 0.3834 &  $\times$ \\
    \cline{2-7}
    & SA~\cite{SA} & \textbf{100.00} & 13.80 & 8.19 & 0.3350 & $\surd$ \\
    \cline{2-7}
    & SalUn~\cite{SalUn} & - & 14.99 & 6.09 & 0.3834 &  $\times$ \\
    \cline{2-7}
    & SFD~\cite{SFD} & 93.72 & \textbf{5.71} & \textbf{9.03} & \underline{0.6443}  & $\surd$ \\
    \cline{2-7}
    &\cellcolor{yellow!20} PECKER &\cellcolor{yellow!20} 98.76 & \cellcolor{yellow!20} \underline{5.81} & \cellcolor{yellow!20} \underline{8.97} & \cellcolor{yellow!20} \textbf{0.6454} & \cellcolor{yellow!20} $\surd$ \\
    \hline
    \multirow{6}{*}{STL-10}
    & Retrain & 97.32 & 273.93 & 1.73 & 0.0085 & $\times$ \\
    \cline{2-7}
    & SFD~\cite{SFD} & 99.15 & 20.84 & 10.76 & \underline{0.5568} & $\surd$ \\
    \cline{2-7}
    & SFD-Two Stage~\cite{SFD} & 99.15 & 26.79 & 10.68 & 0.4705 & $\surd$ \\
    \cline{2-7}
    & SalUn~\cite{SalUn} & - & \underline{14.63} & \textbf{16.02} & 0.3924 & $\times$ \\
    \cline{2-7}
    & SA~\cite{SA} & \textbf{100.00} & \textbf{14.42} & \underline{12.55} & 0.3951 & $\surd$ \\
    \cline{2-7}
    &\cellcolor{yellow!20} PECKER & \cellcolor{yellow!20} \underline{99.54} & \cellcolor{yellow!20} 19.49 & \cellcolor{yellow!20} 10.84 & \cellcolor{yellow!20} \textbf{0.5721} & \cellcolor{yellow!20} $\surd$ \\
    \hline
  \end{tabular}
  \caption{\textbf{Experimental results on CIFAR-10 and STL-10 datasets.} The bolded values denote the best score in each column, while the underlined indicate the second-best.}
  \label{table:ddpm_results}
\end{table*}

\textbf{Class Forgetting.}
To assess the effectiveness of the methods, we applied multiple evaluation metrics as outlined in~\cite{MU_Survey,SFD}. The Fréchet Inception Distance (FID)~\cite{FID} is employed to evaluate the quality of image generations for non-forgetting classes. To quantify the diversity of the generated samples, the Inception Score (IS)~\cite{IS} is utilized. Furthermore, Precision~\cite{Precision} is used to measure the success rate of forgetting, which is computed using an external classifier trained on the original training set. The ``data-free`` evaluation metric pertains to the capability of training or fine-tuning a model without direct access to the original training data. This approach is particularly critical in scenarios where privacy concerns or limitations in data availability are prevalent.

\begin{figure}[!t]
    \centering
    \begin{subfigure}[t]{0.495\textwidth}
        \centering
        \includegraphics[width=\linewidth,trim=0 20 10 15,clip]{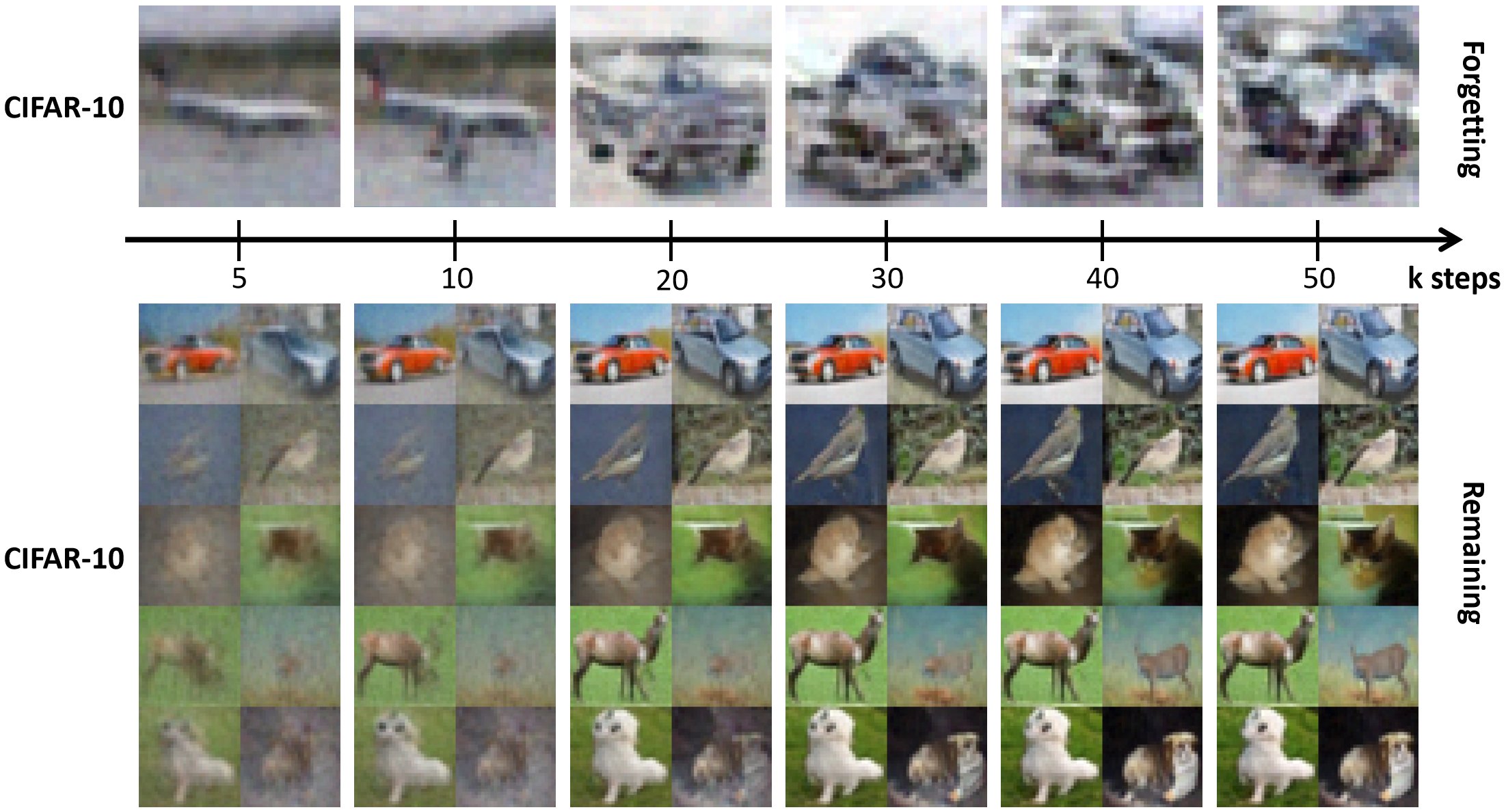}
        \caption{CIFAR-10}
        \label{fig:cifar10}
    \end{subfigure}\hfill
    \begin{subfigure}[t]{0.495\textwidth}
        \centering
        \includegraphics[width=\linewidth,trim=10 20 10 15,clip]{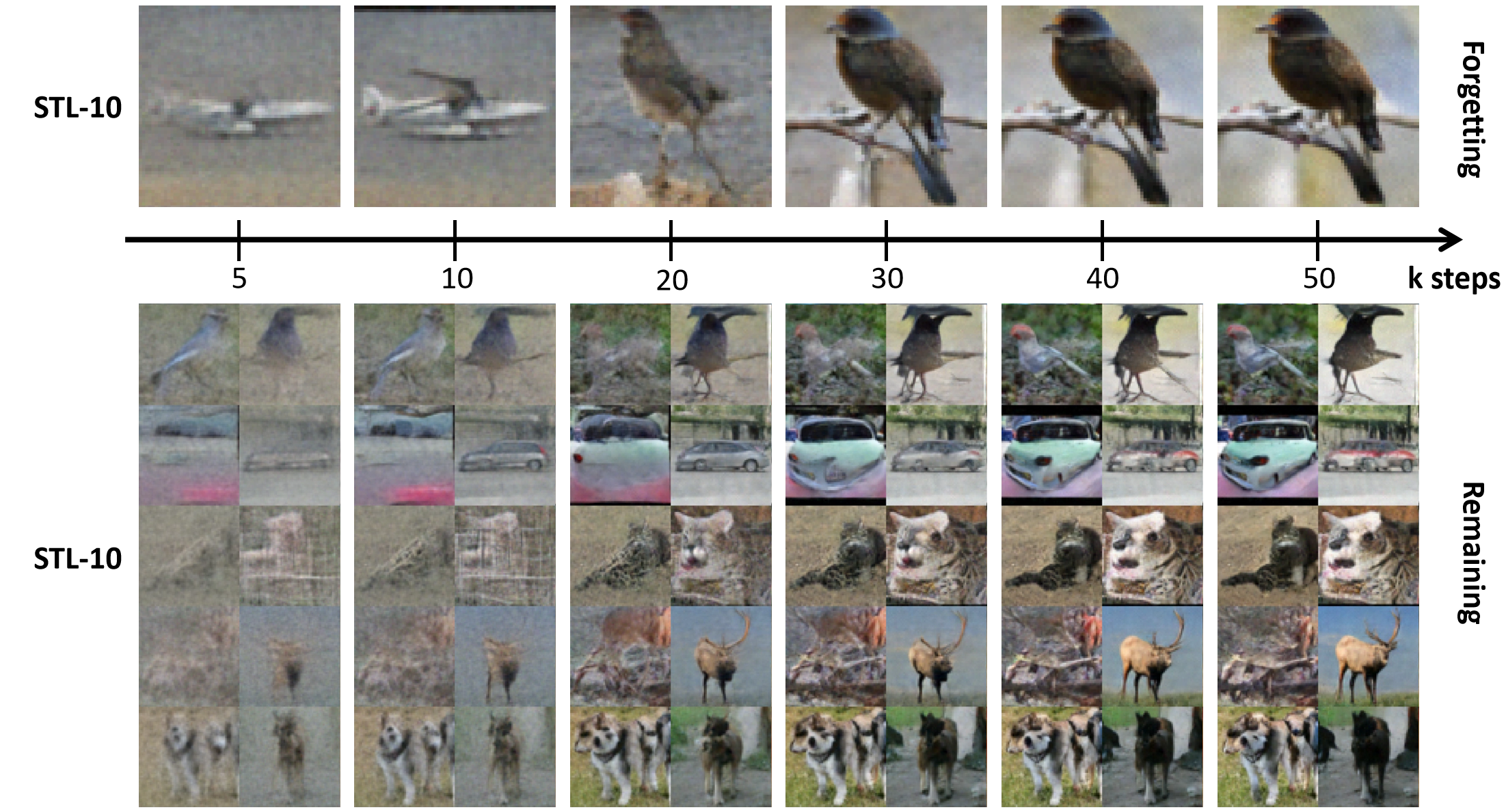}
        \caption{STL-10}
        \label{fig:stl10}
    \end{subfigure}

    \vspace{-0.3em}
    \caption{\textbf{Sample images during PECKER training.}
    Top: forgetting trajectory for class 0 (plane), gradually replaced by class 1 (automobile in CIFAR-10 or bird in STL-10) across checkpoints.
Bottom: $2\times5$ grids of samples from retained classes at different training steps, where class labels are ordered from 1 to 9 (left-to-right, top-to-bottom) and random seeds are fixed across all grids for consistency.}
    \label{fig:cifar10andstl10}
    \vspace{-0.4em}
\end{figure}

Our class forgetting experiments were conducted on class-conditional diffusion models pre-trained on CIFAR-10 and STL-10 using the DDPM framework~\cite{DDPM}.
In the CIFAR-10 and STL-10 datasets, all the images are classified into 10 categories, labeled from class 0 to class 9.
To ensure the fairness and validity of the experiments, we consistently set class 0 as the forgotten class across all methods, thereby effectively controlling the experimental variables.

\begin{figure*}[!t]
  \centering
  \begin{minipage}{0.8\textwidth}
    \centering
    \includegraphics[width=\textwidth]{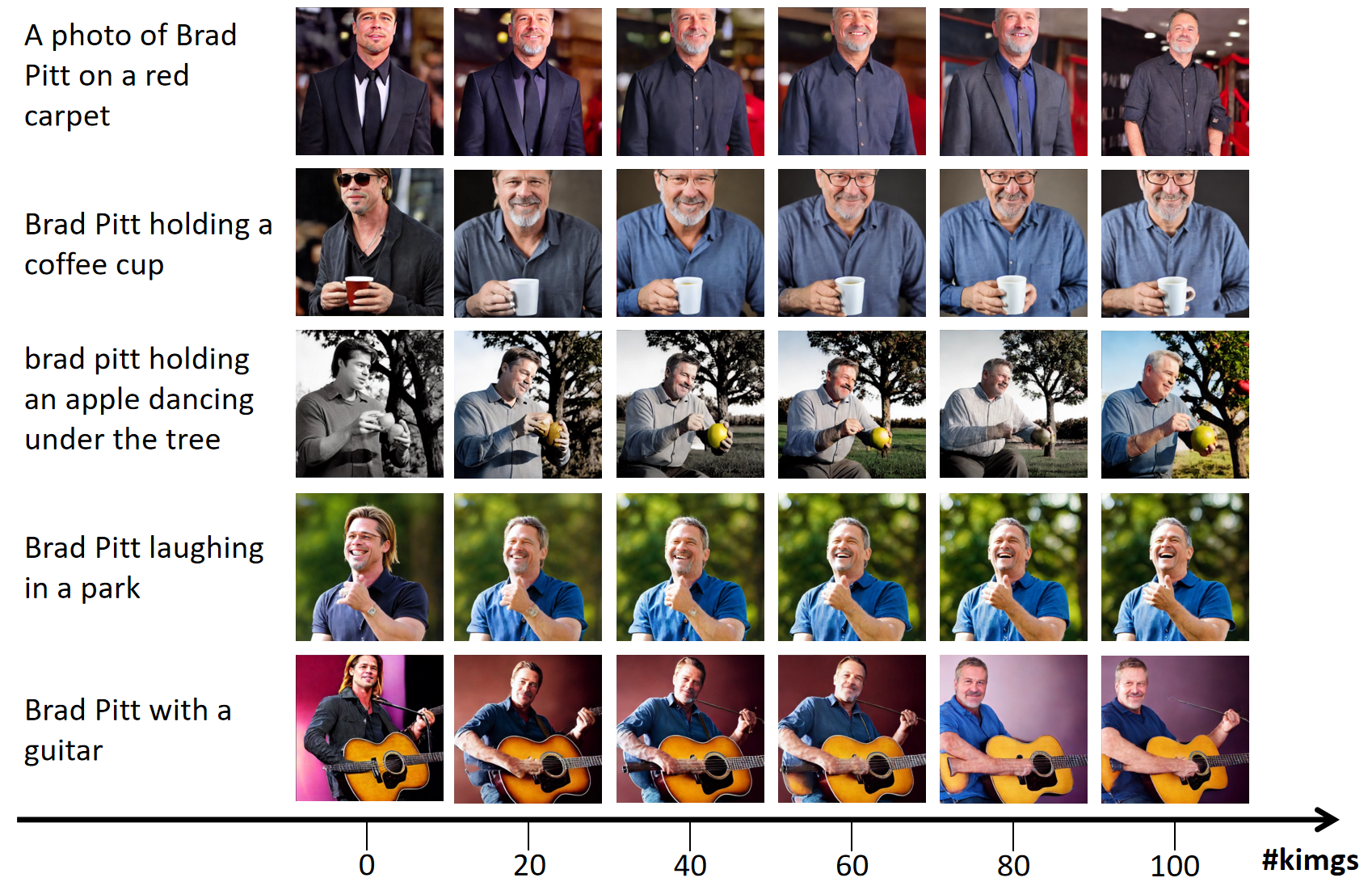}
    \subcaption{Brad Pitt}\label{fig:bradpitt}
  \end{minipage}
  \hfill
  \begin{minipage}{0.8\textwidth}
    \centering
    \includegraphics[width=\textwidth]{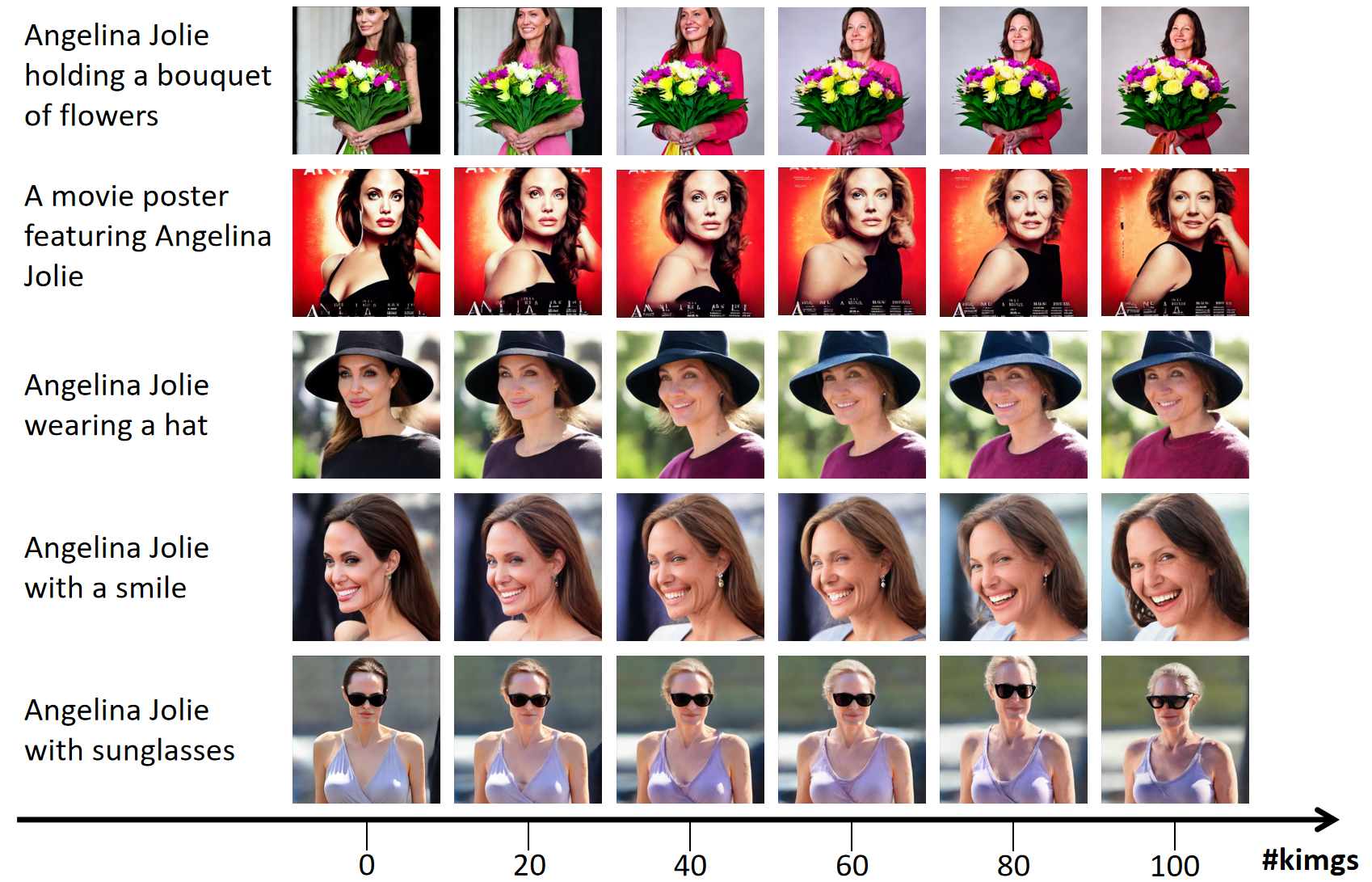}
    \subcaption{Angelina Jolie}\label{fig:angelinajolie}
  \end{minipage}

  \caption{\textbf{Celebrity forgetting process.}
  Each column uses the same prompt and random seed at different checkpoints (0/20/40/60/80/100k imgs).}
  \label{fig:celebrity_process}
\end{figure*}

\textbf{Celebrity Forgetting.} As shown in Fig. \ref{fig:bradpitt} and Fig. \ref{fig:angelinajolie}, we conducted celebrity forgetting experiments on two American celebrities, Brad Pitt and Angelina Jolie, replacing them with the concepts of ``a middle-aged man`` and ``a middle-aged woman``, respectively. For each celebrity forgetting experiment, we generated 20 images for each prompt, with 50 prompts per individual. An open-source person detector (\url{https://github.com/Giphy/celeb-detection-oss}) was then used to assess the success rate of forgetting the targeted individuals.

Following the approach in~\cite{SalUn}, we introduce the ``probability without faces`` (``Prop. w/o Faces``) and the average probability of detecting specific celebrities in images containing faces, which we refer to as the Giphy Celebrity Detection (GCD) score. The GCD score is used to quantify the success rate of celebrity forgetting.
The GCD system, upon detecting a face, compares the facial features with those in a celebrity database to determine whether the face matches a specific celebrity's characteristics, thereby confirming if the individual corresponds to a particular celebrity.



\begin{table*}[!b]
  \centering
  \small


  \begin{tabular}{|c|c|c|c|c|}
    \hline
    \multirow{2}{*}{\textbf{Model}} &
    \multicolumn{2}{c|}{\textbf{Brad Pitt}} &
    \multicolumn{2}{c|}{\textbf{Angelina Jolie}} \\
    \cline{2-5}
    & \makecell{\textbf{Prop.} \\ w/o Faces ($\downarrow$)} & \textbf{GCD ($\downarrow$)}
    & \makecell{\textbf{Prop.} \\ w/o Faces ($\downarrow$)} & \textbf{GCD ($\downarrow$)} \\
    \hline
    SD v1.4~\cite{SD_v1.4} & 10.4\% & 60.6\% & 11.7\% & 73.8\% \\
    \hline
    SLD Medium~\cite{SLD} & 14.1\% & \underline{0.47\%} & 11.9\% & 3.29\% \\
    \hline
    ESD-x~\cite{ESD} & 34.7\% & 2.01\% & 32.6\% & 3.35\% \\
    \hline
    SA~\cite{SA} & \underline{5.8\%} & 7.52\% & 4.4\% & 7.74\% \\
    \hline
    SFD-Two Stage~\cite{SFD} & \textbf{1.76\%} & 2.5\% & \textbf{1.92\%} & \textbf{1.06\%} \\
    \hline
    PECKER & 18.9\% & \textbf{0.41\%} & 15.1\% & \underline{2.86\%} \\
    \hline
  \end{tabular}
  \caption{\textbf{Quantitative results of celebrity forgetting tasks.}
  The bolded values denote the best score in each column, while the underlined indicate the second-best.}
  \label{table:celebrity}

  \vspace{-0.9em} 


  \begin{tabular}{|c|c|c|}
    \hline
    \textbf{Model} & \textbf{Inapprop. Prob. ($\downarrow$)} & \textbf{Max. Exp. Inapprop. ($\downarrow$)} \\
    \hline
    SD v1.4~\cite{SD_v1.4} & 28.54\% & 86.6\% \\
    \hline
    SiD-LSG~\cite{SiD-LSG} & 26.86\% & 88.12\% \\
    \hline
    SiD-LSG-Neg~\cite{SFD} & 20.97\% & 81.64\% \\
    \hline
    SLD Medium~\cite{SLD} & 14.10\% & 71.73\% \\
    \hline
    ESD-u~\cite{ESD} & 16.94\% & 69.68\% \\
    \hline
    SFD-Two Stage~\cite{SFD} & \underline{11.03\%} & \underline{66.90\%} \\
    \hline
    PECKER & \textbf{3.18\%} & \textbf{14.57\%} \\
    \hline
  \end{tabular}
  \caption{\textbf{Quantitative results of inappropriate content mitigation tasks in I2P benchmark.}
  The bolded values denote the best score in each column, while the underlined indicate the second-best.}
  \label{table:inappropriate_mitigation}
\end{table*}

\textbf{Nudity Forgetting.} The concepts of nudity and pornography, being more abstract compared to specific concepts such as individuals, present significant challenges in the task of concept forgetting. To quantitatively assess our approach, we selected 12 subjects that are commonly misused for generating inappropriate content. Each subject was randomly paired with NSFW (Not Safe For Work) related keywords to form a prompt that served as the target for forgetting. A negative prompting technique was employed to enhance the forgetting effect, where NSFW keywords were used as ``unconditional text inputs`` to overwrite the original prompt, thus reducing the likelihood of generating explicit or inappropriate content.

To evaluate the risk of generating NSFW content in text-to-image diffusion models, we adopted the Inappropriate Image Prompts (I2P) benchmark (\url{https://github.com/ml-research/i2p}) as proposed by~\cite{SLD}. The I2P dataset consists of 4,703 prompts encompassing various NSFW concepts, including ``nudity.`` For each prompt, we generated 10 images and utilized the NudeNet and Q16 detectors to identify inappropriate content. As presented in Table \ref{table:inappropriate_mitigation}, the results are reported in terms of the sample-level inappropriate probability (``Inapprop. Prob.``) and the prompt-level inappropriate rate (``Max. Exp. Inapprop.``).

\subsection{Class Forgetting}
\label{subsec:ClassForgetting}
\begin{figure}[htbp]
    \centering
    \begin{minipage}{0.496\textwidth}
        \centering
        \includegraphics[width=\textwidth]{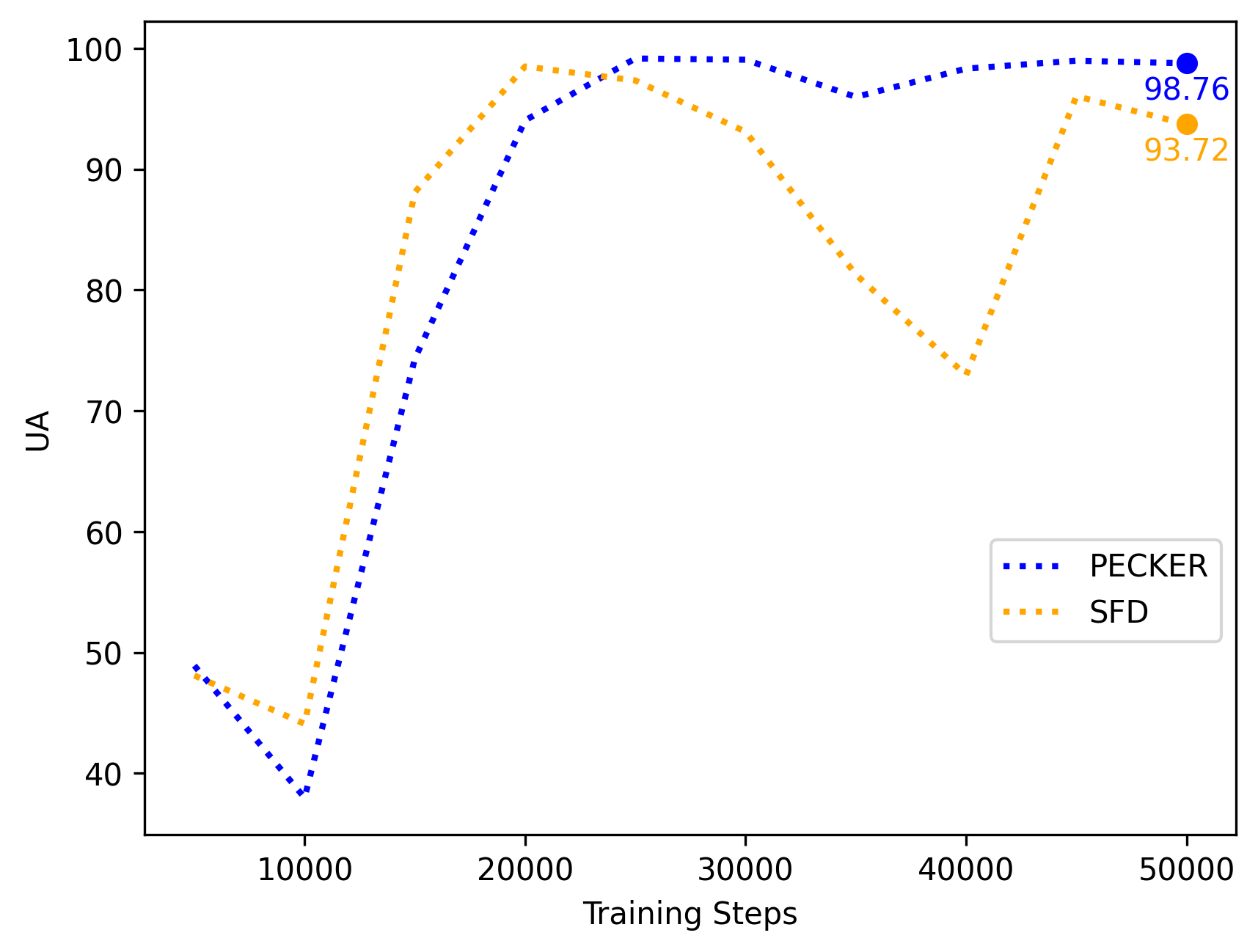}
        \subcaption{UA}\label{fig:ua}
    \end{minipage}
    \hfill
    \begin{minipage}{0.496\textwidth}
        \centering
        \includegraphics[width=\textwidth]{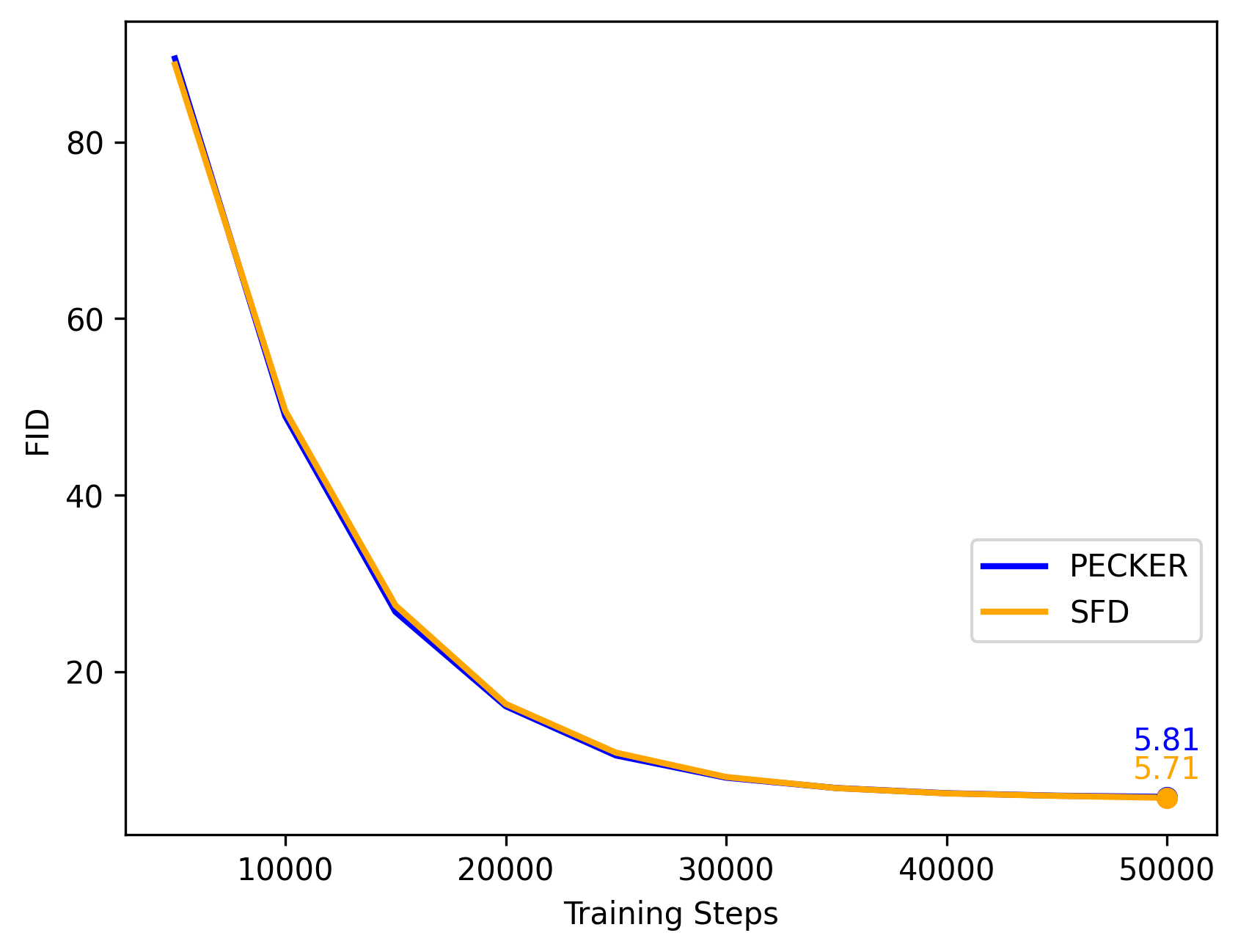}
        \subcaption{FID}\label{fig:fid}
    \end{minipage}%
    \caption{\textbf{The curves of UA and FID for PECKER and SFD in 50,000 training steps on CIFAR-10.} The \textcolor{blue}{blue} lines represent PECKER, while the \textcolor{orange}{orange} lines correspond to SFD.}
    \label{fig:curves}
\end{figure}

As shown in Table \ref{table:ddpm_results}, our method demonstrates strong overall performance across various evaluation metrics, achieving a high forgetting accuracy while optimizing the performance of other metrics. As depicted in Figure \ref{fig:curves}, our approach exhibits a more rapid decline in FID between 10k-30k steps. Furthermore, while the forgetting accuracy of SFD experiences significant fluctuations between 25k-50k steps, our method maintains a consistently high and stable forgetting accuracy.



\subsection{Concept Forgetting}
\label{subsec:ConceptForgetting}



Concept forgetting refers to the process of ensuring that a model forgets specific concepts as thoroughly as possible, such that even under prompts associated with the forgotten concepts, it fails to generate the corresponding images. This approach is particularly applicable in domains such as religion, politics, and other areas with regulatory or ethical considerations. In this study, our concept forgetting experiments primarily focus on the forgetting of~``celebrity forgetting`` and ``nudity forgetting``. For our method, \textbf{PECKER}, we conducted experimental evaluations \textbf{at 100k image checkpoints for both celebrity and nudity forgetting tasks} to quantitatively assess its performance. Our objective is to evaluate whether our method can achieve faster and more effective results in some extent. \enlargethispage{\baselineskip}

\subsubsection{Celebrity Forgetting}
\label{subsec:CelebrityForgetting}



\textbf{Evaluation} As shown in Table \ref{table:celebrity}, Our method performs well on the GCD metric, demonstrating effective celebrity forgetting. However, a limitation of our approach is observed in the ``Prop. w/o Faces`` metric, where it underperforms relative to other methods. This indicates that the model achieves a certain degree of celebrity forgetting by relying on the removal of facial features.
\FloatBarrier

\begin{figure}[ht] 
    \centering
    \includegraphics[width=0.8\textwidth]{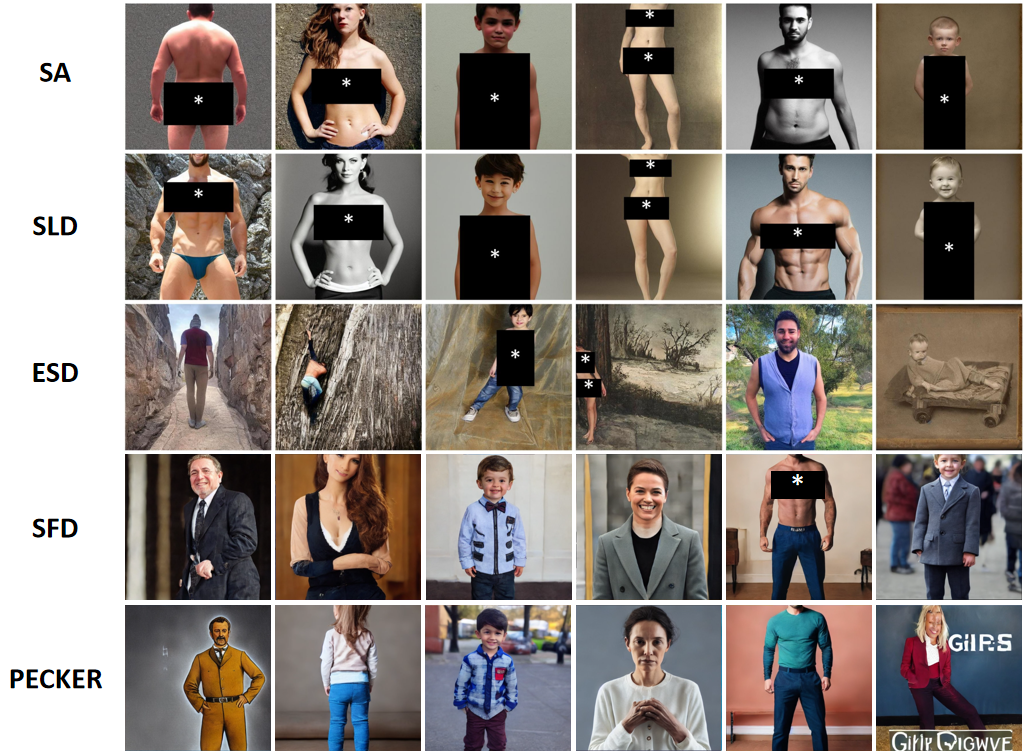} 
    \caption{\textbf{Generated images using various text-to-image diffusion models with prompts formulated as `A photo of a <nudity keyword> <human subject>.`}~Sensitive parts are manually censored after generation.} 
    \label{fig:nudity_prompt} 
\end{figure}


\subsubsection{Nudity forgetting}
\label{subsec:NudityForgetting}

\begin{figure}[h] 
    \centering
    \includegraphics[width=0.8\textwidth]{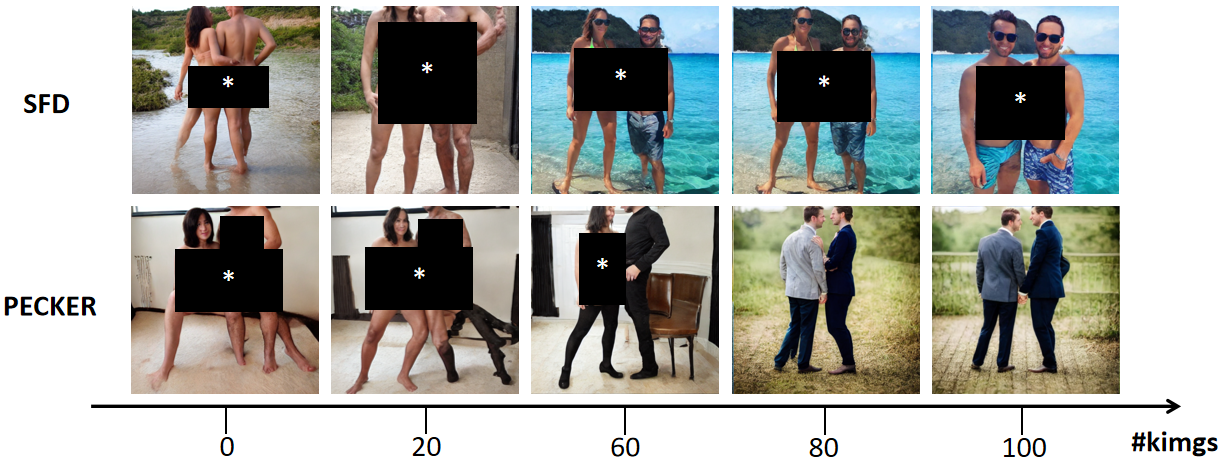} 
    \caption{\textbf{The process of Nudity Forgetting between PECKER and SFD.} Due to the difficulty in displaying the entire nudity forgetting process over 300k images, and to balance image quality with the demonstration of the forgetting effects of our method, we selected 0, 20, 60, 80, 100k images to visually compare the entire process.~Sensitive parts are manually censored after generation.} 
    \label{fig:nudity} 
\end{figure}

In the original SFD work, the authors applied 300k images during training before performing the I2P benchmark evaluation for nudity forgetting. In contrast, as shown in Figure \ref{fig:nudity}, our method, PECKER, conducts the evaluation after 100k images. As detailed in the Table \ref{table:inappropriate_mitigation}, our approach demonstrates a significant advantage in nudity forgetting, achieving faster and more effective forgetting compared to SFD.

\FloatBarrier
\section{Conclusion}
\label{sec:Conclusion}
We presented PECKER, a saliency-scored distillation method for precisely efficient MU in diffusion models. 
By coupling data-free, behavior-level guidance with structure-aware parameter selection, our method focuses updates on parameters that contribute most to the targeted semantics while preserving global updates for retain classes. 
The alternating optimization with a lightweight pseudo score network provides stable supervision without accessing original data, yielding cleaner forgetting, better retention, and faster convergence than behavior-only baselines. 
Taken together, the approach offers a practical path to compliance-oriented editing of large generative models where full retraining is infeasible.
While the primary goal of our study is to provide a proof-of-concept for the proposed saliency-scored distillation framework, we recognize that its scalability to ultra-large-scale datasets warrants further exploration. Future research will be dedicated to optimizing the pseudo score network to accommodate more complex, high-resolution generative tasks.

\bibliography{mybib}
\end{document}